\documentclass{article}
\usepackage[nonatbib, preprint]{neurips_2019}
\usepackage[utf8]{inputenc} 
\usepackage[T1]{fontenc}    
\usepackage{hyperref}       
\usepackage{url}            
\usepackage{float}
\usepackage{capt-of}
\usepackage{booktabs}       
\usepackage{amsfonts}       
\usepackage{nicefrac}       
\usepackage{microtype}      
\usepackage{graphicx}
\usepackage{xcolor}
\graphicspath{{Figures/}}

\title{Improved object recognition using neural networks trained to mimic the brain’s statistical properties}

\author{
  Callie Federer\\
  Department of Physiology and Biophysics\\
  University of Colorado Anschutz Medical Campus \\
  Aurora, CO \\ 
  \texttt{callie.federer@ucdenver.edu} \\
  \And
   Haoyan Xu \\
   Department of Computing Science \\
   University of Alberta \\
   Edmonton, AB \\
   \texttt{haoyan5@ualberta.ca} \\
  \And
   Alona Fyshe\\
   Department of Computing Science \\ 
   University of Alberta \\
   Edmonton, AB \\
   \texttt{alona@ualberta.ca} \\
  \And
   Joel Zylberberg \\\emph{•}
   Learning in Machines and Brains Program \\ 
   Canadian Institute For Advanced Research (CIFAR)  \\
   Toronto, ON \\
  \texttt{joelzy@yorku.ca} \\
}

\begin{document}

\maketitle

\begin{abstract}
The current state-of-the-art object recognition algorithms, deep convolutional neural networks (DCNNs), are inspired by the architecture of the mammalian visual system, and are capable of human-level performance on many tasks. As they are trained for object recognition tasks, it has been shown that DCNNs develop hidden representations that resemble those observed in the mammalian visual system~\cite{Kriegeskorte2014, Yamins2016, Gu2015, Mcclure2016}. Moreover, DCNNs trained on object recognition tasks are currently among the best models we have of the mammalian visual system. This led us to hypothesize that teaching DCNNs to achieve even more brain-like representations could improve their performance. To test this, we trained DCNNs on a composite task, wherein networks were trained to: a) classify images of objects; while b) having intermediate representations that resemble those observed in neural recordings from monkey visual cortex. Compared with DCNNs trained purely for object categorization, DCNNs trained on the composite task had better object recognition performance and are more robust to label corruption. Interestingly, we found that neural data was not required for this process, but randomized data with the same statistical properties as neural data also boosted performance. While the performance gains we observed when training on the composite task vs the ``pure" object recognition task were modest, they were remarkably robust. Notably, we observed these performance gains across all network variations we studied, including: smaller (CORNet-Z) vs larger (VGG-16) architectures; variations in optimizers (Adam vs gradient descent); variations in activation function (ReLU vs ELU); and variations in network initialization. Our results demonstrate the potential utility of a new approach to training object recognition networks, using strategies in which the brain -- or at least the statistical properties of its activation patterns -- serves as a teacher signal for training DCNNs.
\end{abstract}

\section{Introduction}
Deep convolutional neural networks (DCNNs) have recently led to a rapid advance in the state-of-the-art object recognition systems \cite{Lecun2015}. At the same time, there remain critical shortcomings in these systems~\cite{Rajalingham2018}. We asked whether training DCNNs to respond to images in a more brain-like manner could lead to better performance. DCNN architectures are directly inspired by that of the mammalian visual system (MVS) \cite{Hubel1968}, and as DCNNs improve at object recognition tasks, they learn representations that are increasingly similar to those found in the MVS \cite{Kriegeskorte2014, Yamins2016, Gu2015, Mcclure2016}. Consequently, we expected that forcing the DCNNs to have image representations that were \emph{even more} similar to those found in the MVS, could lead to better performance.
  
Previous work showed that the performance of smaller ``student" DCNNs could be improved by training them to match the image representations of larger ``teacher" DCNNs ~\cite{Mcclure2016, FitNets2015, Hinton2015}, and that DCNNs could be directly trained to reproduce image representations formed by the V1 area of monkey visual cortex \cite{Kindel2017}. These studies provide a foundation for the current work, in which we used monkey V1 as a teacher network for training DCNNs to categorize images. We then tested the hypothesis that  DCNNs trained with the monkey V1 as a teacher would outperform those trained without this teacher signal. By several relevant metrics (including accuracy), we found that performance was increased when monkey V1 was used as a teacher. Importantly, the monkey V1 data were collected in response to different images than the ones in the object recognition task. As a result, our approach, of using the brain as a teacher signal, can leverage pre-existing, and publicly-available neural data, without necessarily requiring new neuroscience experiments for each new machine learning task. Moreover, we also trained DCNNs with random teacher signals that matched the statistics of monkey V1 neural activations, and found that those outperformed networks trained without a teacher signal. This emphasizes a potential role for using the statistical properties of neural activations as a form of regularizer, that could be useful for training DCNNs.

Related recent work has demonstrated success in using neural data to train machine learning models. One study found that using fMRI measurements of human brain activations from subjects viewing images could guide Support Vector Machines (SVMs) decision boundaries \cite{Fong2018}. In this study, the authors weighted the training data based on  how easy it was for the human brain to recognize the example as a member of a class \cite{Fong2018}. This is different from our work where we train deep convolutional neural networks with a two-part cost function that explicitly trains on matching the \emph{neural representations}, as opposed to the previous approach which weighted the cost function on specific training examples. In our work, the images shown to the animal during the collection of neural data were not category-labelled images from a machine learning benchmark task. For contrast, in the recent Fong et al. study \cite{Fong2018}, the neural data had to be collected for the same image set used for the categorization task.  Previous work from Peterson \textit{et al. } demonstrated that training deep convolutional neural networks with human perceptual uncertainty makes classification more robust to variations in the test set and to adversarial examples \cite{Peterson2020}. In this study, they use human guidance to change the labels for training, incorporating uncertainties. We instead focus on changing the cost function -- incorporating neural data into the network evaluation -- without considering behavioral reports of uncertainty. Finally, previous work from Linsley et al. \cite{Linsley2019} demonstrated that using human behavioral data to add supervisory attention guidance improved object recognition performance. This is again quite a different approach from ours: while they focused on behavioral data, we instead used signals recorded from visual cortical neurons. 

Notably, we did not aim to achieve state-of-the-art classification performance in this work: we instead sought to test whether the use of neural data as a "teacher" signal could robustly improve DCNN performance. For that reason, we studied a wide variety of network properties: different architectures and network sizes; different activation functions; and different optimizers. Our results indicate that, over all of these variations,  \textbf{(1)} DCNNs trained to mimic monkey V1 (or surrogate data matching the statistics of monkey V1) have better object recognition performance, \textbf{(2)} DCNNs trained to mimic monkey V1 make fewer errors, and of these errors, more of them are within the correct superclass, and \textbf{(3)} DCNNs trained to mimic monkey V1 are more robust to label corruption. While the performance gains we observed were somewhat modest, based on the robustness of those performance gains, we anticipate that future work could productively apply our new training method to other networks, potentially improving on the current state-of-the-art object recognition systems.

\section{Methods}
\subsection{Monkey Visual Cortex Data}
Our monkey V1 teacher signal is from publicly-available multielectrode recordings from anesthetized monkeys presented with a series of images while experimenters recorded the spiking activity of neurons in primary visual cortex (V1) with a multielectrode array \cite{Coen-cagli2015}. These recordings were conducted in 10 experimental sessions with 3 different animals, resulting in recordings from 392 neurons. The monkeys were shown 270 static natural images as well as various static grating images for 100 ms presentations. A sample of these images can be found in Fig. \ref{sample_figs}: we used the responses to the natural image stimuli for our experiments, and did not use the responses to the grating stimuli. 

In addition to these V1 data, which are our main focus, we also studied recordings from cortical areas V4 and IT \cite{Majaj2015}. Similar to the V1 data, these recordings were performed with multielectrode arrays implanted in the cortex, and neural responses were recorded while the animals viewed images. Importantly, the images shown to the animals in these experiments are not from the CIFAR-100 dataset (nor any machine learning benchmark dataset). Thus, our approach is not limited to cases in which the neural data is recorded for the same stimulus set that defines the machine learning task.
\subsection{Representational Similarity Matrices (RSMs)}
To compare image representations in the monkey brain with those in a DCNN, we used representational similarity matrices (RSMs) as the teacher signal \cite{Kriegeskorte2008}. For each pair of images (i \& j) shown to the monkey, we computed the similarity between measured neural responses  ($v_{i}$ \& $v_{j}$): these vectors contain the firing rates of all of the observed neurons. We measured the similarity using the cosine similarity between those vectors  
\begin{equation}
	RSM_{ij} =  \frac{v_{i} \cdot v_{j}}{\mid \mid v_{i} \mid \mid \times \mid \mid v_{j} \mid \mid}
\end{equation}

where $v_{i}$ is the neural response to image i and $v_{j}$ is the neural response to image j.

These values ($RSM_{ij}$) were assembled into matrices, describing the representational similarities \cite{Kriegeskorte2008} in monkey V1, for all image pairs (i \& j). We averaged the representational similarity matrices over the 10 experimental sessions to yield a single RSM that was used for training the neural networks. During DCNN training, we input the same pairs of images (i \& j) into our DCNNs as were displayed to the monkeys, and computed representational similarity matrices for the chosen layer of hidden units (Fig. 1): the DCNN's RSM is denoted by $\widehat{RSM}$.
\subsection{Deep Convolutional Neural Networks and Cost Functions}
We performed most of our experiments on the CORNet-Z DCNN architecture, described below. We also performed some experiments on the much larger VGG-16 architecture. Results were similar for both architectures. The CORNet-Z DCNN architecture \cite{Kubilius2018} is a trimmed-down version of the AlexNet \cite{Krizhevsky2015} object recognition algorithm (Fig. 1). The layers in CORNet-Z have been identified with the brain areas at the corresponding depths within the mammalian visual hierarchy \cite{Kubilius2018} (Fig. 1). 
\begin{figure}[ht]
  \centering
  \includegraphics[trim={0 0cm 0 0}, clip, scale=1]{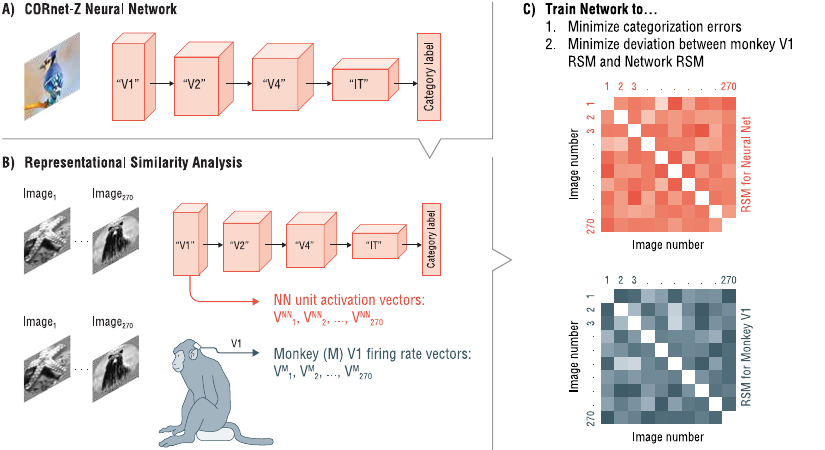}
  \caption{Overview of the algorithm and experiments. We did the majority of our experiments using the CORNet-Z architecture; some of our experiments were also done using the larger VGG-16 architecture (not shown). A) The architecture of CORNet-Z, which has multiple blocks. Each block consists of a convolution followed by a ReLU nonlinearity and max pooling. We found similar results using ELU nonlinearities (Fig. \ref{elu}). The blocks are identified with cortical areas V1, V2, V4 and IT, which exist at the corresponding depths in the primate visual system. B) To compare the neural network's image representation to that of monkey V1, we used a set of 270 images that were shown to monkeys while neural firing rates were recording using an implanted Utah array ($V^M_{1 \dots 270}$). We input the same 270 images into the artificial neural network (top), and extracted the vectors of unit activations in the network's V1 layer ($V^{NN}_{1 \dots 270}$). We then computed the representation similarity matrix (RSM) for the neural network; this 270 x 270 matrix defines how similar the unit activation patterns in the network are, for each image pair. We also computed the RSM for monkey V1, over the same set of 270 images (bottom). C) We trained the neural networks to minimize errors in categorizing CIFAR100 images, while also minimizing the differences between the RSM for the Neural Network RSMs (red), and the RSM for monkey V1 (blue).}
\end{figure}
We trained the networks on the CIFAR100 \cite{Cif100} task, which consists of classifying the objects depicted in images into one of 100 different categories. Regardless of the network architecture, we randomly initialized all weights with the Glorot uniform initializer \cite{Glorot}, and trained the DCNN to minimize a cost function consisting of two terms: classification error, and mismatch between the network's hidden representations and those in monkey V1. Classification error was computed as the cross entropy between the network's final outputs and the true object labels. Representation mismatch was computed as the mean-squared error between the monkey V1 representational similarity matrix (RSM), and that of the relevant layer of the DCNN ($\widehat{RSM}$). For CORNet-Z, this was the V1 block (Fig. 1), and for VGG-16, this was the third convolutional layer. Previous work has shown that, in VGG-16 networks trained for object recognition tasks, the third layer has representations most similar to those seen in monkey V1~\cite{Cadena2017}.

A trade-off parameter, $\lambda$, determines the relative weighting of the two terms in the cost function
\begin{equation}
	cost =  \lambda  \sum_{i,j}(RSM_{ij} - \widehat{RSM}_{ij})^2 - \sum_{i}\hat{y_i}log(y_i) 
\end{equation}
where $\hat{y_i}$ is the network approximation of the output and $y_i$ is the target output. We updated the trade-off parameter $\lambda$ throughout training, so as to keep the ratio between the two terms in the loss function constant. In other words, $\lambda$ was updated so that $r$ was constant, with $r =  \lambda \left[\sum_{i,j}(RSM_{ij} - \widehat{RSM}_{ij})^2 \right]/\left[\sum_{i}\hat{y_i}log(y_i)\right] $. We studied networks with several different values of this ratio, $r$. We also experimented with using a constant $\lambda$ throughout training but found this constant-ratio method leads to better object recognition performance (Fig. \ref{static_ratios}). This cost function is similar to the cost function used in \cite{Kietzmann2019}, where representational dissimilarity analysis was used to compare how well recurrent and feed-forward deep convolutional neural networks could capture human MEG recordings while viewing natural object stimuli. 

\subsection{Training Procedures}
We trained each CORNet-Z network for 100 epochs (250 for VGG-16 , discussed in sec. \ref{VGG}). Networks trained with neural data (i.e., with $r>0$), were trained to minimize the composite cost (Eq. 1) for the first 10 epochs, and thereafter were trained on just the cross entropy loss (this increased from 10 to 100 epochs for the VGG-16 experiments).  In other words, we set $r = 0$ after these first 10 epochs, meaning there was no longer any impact from the monkey V1 ``teacher signal''. This procedure reduces the computational cost due to the need for two separate forward passes, one with the natural images from the monkey experiments and one with the CIFAR100 data. We performed some experiments in which the neural data training signal was applied at all training epochs, and saw similar results (Fig. \ref{allepochs}). 

We kept a static training rate of 0.01 for all networks and a batch size of 128 for CORNet-Z or 256 for VGG-16. Training with an alternative learning rate schedule using the Adam optimizer led to similar conclusions: performance was improved by using neural data in the training procedure (Fig. \ref{adam}). We used dropout regularization \cite{dropout} with 0.5 retention probability for the 3 fully connected layers of all networks. For the training images, we centered the pixels globally across RGB channels. The held-out testing images were not preprocessed, to ensure fair evaluation. The natural images that were presented to the monkeys in the neuroscience experiments were also not preprocessed. The code associated with this paper can be downloaded at github.com/cfederer/TrainCNNsWithNeuralData. 

For each architecture and choice of parameters, we repeated the training from 10 different random initial conditions; results reported are mean $\pm$ SEM. This approach has a larger computational cost than does reporting the result of a single training run, but makes it more likely that our findings will generalize because they do not depend on the idiosyncrasies of weight initializations \cite{Mehrer2020}.
\subsection{Control Experiments with Randomly Generated RSMs}
For control experiments, we tested whether randomly generated RSMs would have the same benefit in object recognition performance as the monkey V1 RSMs. We repeated our experiments with randomly-generated RSMs in place of the monkey V1 ones. We generated the random RSMs in several different ways.

First, we drew 39-element i.i.d. random vectors from a Gaussian distribution with the same mean and variance as were seen in the monkey data: $\mu$ = 0.495 and $\sigma$=0.582; different random vectors were drawn for each image. We then used these random vectors to generate a RSM, as described above. This RSM is referred to as ``Gaussian (V1-stats)". The number of elements (39) matches the average number of neurons simultaneously observed in the monkey experiments~\cite{Coen-cagli2015}.

To test whether matching the statistics of the neural data matters for training the networks, we next drew 39-element vectors from i.i.d. Gaussian distributions with $\mu$ = 5 and $\sigma$ = 0.582, and calculated the representational similarity matrix from these randomly drawn vectors. This RSM is referred to as ``Gaussian (non V1-stats)" because its mean (and mean-variance relationship) differ substantially from the neural data.

Finally, we applied a shuffling procedure to the V1 data, where we randomly permuted the image identities associated with each recorded vector of neural firing rates. As a result of this procedure, the neural responses no longer matched the images that were shown to the monkey. The random permutation was done independently for each neuron. This leads to vectors of firing rates that match (for each neuron, and to all orders) the distributions seen in the monkey data, but removes information about the specific image features those neurons represent (i.e., the neurons' receptive fields). Similar to the above experiments, we assembled these vectors into a RSM. This RSM is referred to as ``V1 shuffled".

\section{Results}
We trained neural networks on the composite cost (Eq. 1), with varying ratios $r$ describing the trade-off between representational similarity cost and categorization cost. We evaluated the trained networks based on categorization accuracy achieved on held-out data (not used in training) from the CIFAR100 dataset. We ran each experiment 10 times with different initial randomized conditions to demonstrate that the difference in accuracy is not due to initial conditions \cite{Mehrer2020}.  In our figures, black lines indicate networks trained purely for categorization ($r=0$), while red lines indicate networks trained using monkey V1 as a teacher ($r>0$; see Methods). Higher weightings of neural data in the loss function correspond to lighter red lines.
\subsection{Networks trained to Mimic Monkey V1 Image Representations Have Higher Object Recognition Accuracy}
We first present results from the CORNet-Z architecture, and we discuss results from the larger VGG-16 model in sec. \ref{VGG}. We tested the trained models' ability to classify previously-unseen images from the CIFAR100 dataset (i.e., images not used in training), and quantified the fraction of images correctly labeled. Networks trained to both classify objects and match neural representations (i.e., those with $r>0$) have better object recognition performance than those trained without using the monkey brain as a teacher (i.e., those with r=0; Fig. 2). The effect is not monotonic: maximum object recognition performance is achieved with $r=0.1$; larger values of $r$ lead to a reduction in object recognition performance, presumably because the training procedure places an insufficient emphasis on categorization (e.g., cross entropy). The distributions of test set accuracy, over the 10 different initializations, for networks trained with $r=0$ and $r=0.1$, can be found in Fig. \ref{hist}. There is a clear separation in these distributions, highlighting that the performance gain obtained by using neural data in the training procedure is robust against variations in initialization.
\begin{figure}[t]
  \centering
  \includegraphics[trim={0 7cm 0 0}, clip, scale=.47]{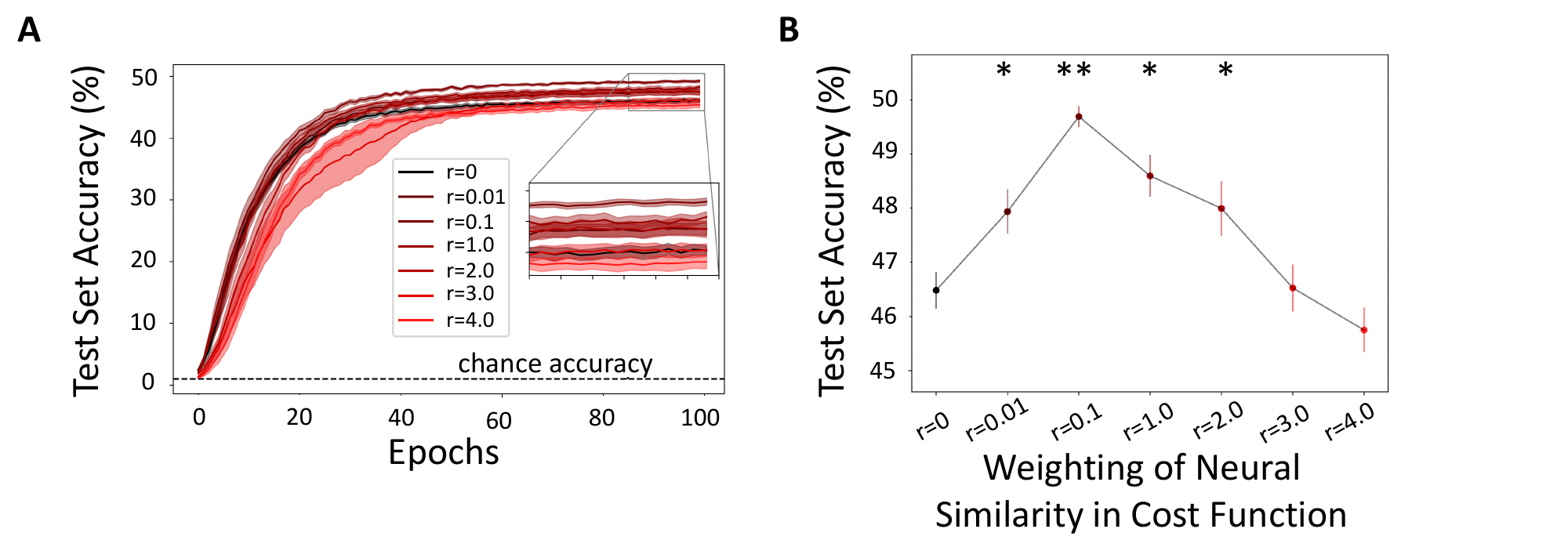}
  \caption{Accuracy in categorizing previously-unseen CIFAR100 images for the CORNet-Z architecture trained with different weighting ratios, $r$, applied to monkey V1 representation similarity for the first 10 epochs of training. A) Test-set accuracy at each epoch during training. Chance accuracy is indicated by the dashed black line. Shaded areas are +/- SEM over 10 different random initializations of each model.  B) Test-set accuracy for previously-unseen CIFAR100 images at the end of training, as a function of the weight ($r$) given to neural representational similarity in the cost function. Double asterisks (**) indicate significantly higher results from no neural data, $r=0$, at p<.001 on a one-tailed t-test. Single asterisks (*) indicate significantly higher results from no neural data, $r=0$, at p<.05 on a one-tailed t-test. }
\end{figure}
\subsection{Networks trained to Mimic Monkey V1 Image Representations Have More Diverse Unit Activations}
How do the representations in the V1-like layers differ between networks trained with, or without, neural data?  To gain insight into this question, we used t-SNE \cite{Maaten2008} to visualize the unit activations from the first convolutional layer of the networks from the preceding section (i.e., the layer of CORNet-Z that aligns with monkey V1 in terms of depth in the visual pathway). We input images from the CIFAR100 test set into the networks, and used t-SNE to embed those network activations into two dimensions. We repeated this procedure for networks trained with no monkey V1 teacher signal, $r=0$, and with monkey V1 teacher signal, $r=0.1$. 

While the representations of images from different categories remain co-mingled at this low level of the neural network, the representations are more varied for the network trained with a neural representation weighting of $r=0.1$ (Fig. 3b) than for the network trained no neural data ($r=0$: Fig. 3a). This motivated us to compute the average variance per unit within the V1-like layer of the CORNet-Z networks (variance in activations over the test-set images, averaged over all units in that hidden layer). As the weighting ratio $r$ for neural similarity in the cost function increases, so too does the activation variance (Fig. 3c). While this increase in activation variance is initially associated with increasing object-recognition performance (e.g., up to $r=0.1$), at higher values of $r$, that increased unit activation variance no longer correlates with higher accuracy (e.g., for $r=3.0$ or $r=4.0$: Fig. 3c). These data suggest that the improved generalization performance obtained by using neural data in the training procedure (i.e., Fig. 2) could arise in part because the training procedure that uses neural data forces the networks to have more diverse activations in their low-level units. Variance in activations alone does not explain the increase in categorization performance: while variance keeps increasing for $r$ values beyond $0.1$, categorization performance peaks for $r=0.1$ and declines for larger values of $r$.
\begin{figure}[t]
  \centering
  \includegraphics[trim={0 7cm 0 0}, clip, scale=.42]{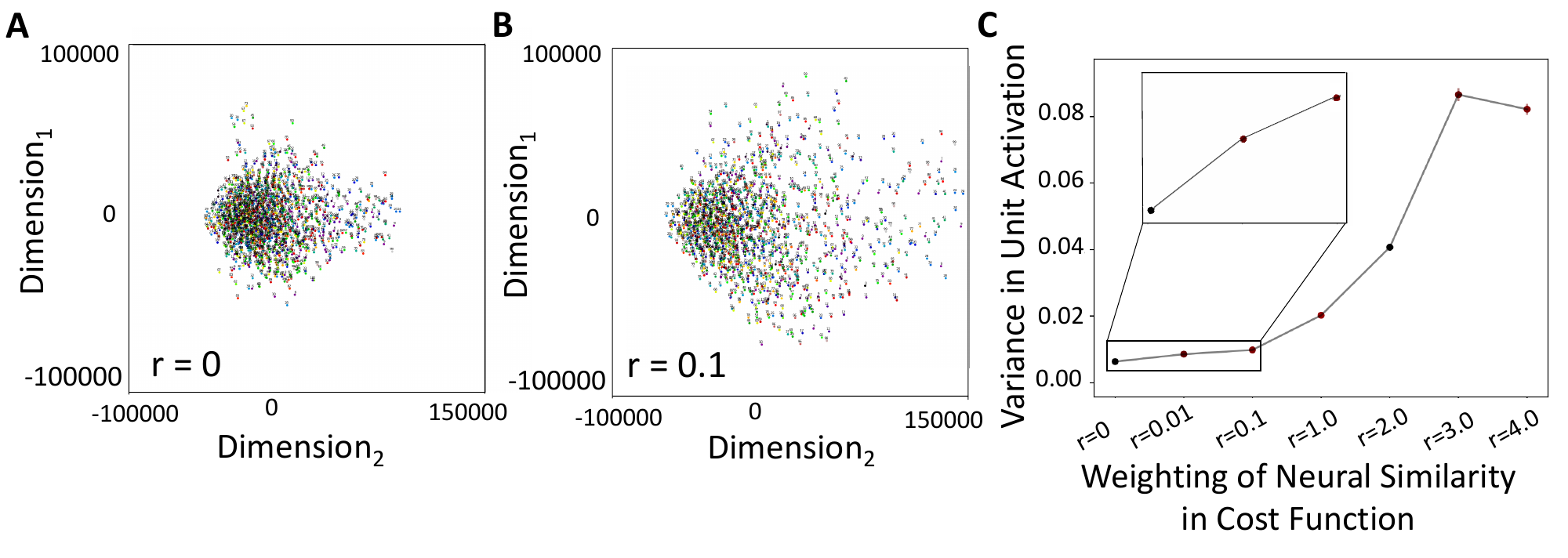}
  \caption{Visualizing the hidden unit activations of the first convolutional layer, which is the V1-like layer of CORNet-Z. In panels A and B, we input the same 1280 randomly-selected images from the CIFAR100 test set into DCNNs trained with either $r=0$ or $r=0.1$. We then used t-SNE to embed those high-dimensional unit activations into two dimensions.  X- and Y-axes represent the two dimensions of this t-SNE embedding. Colors indicate the object categories for each input image. A) t-SNE embedding of hidden-unit activations from networks trained with no neural data ($r=0$).  B)  t-SNE embedding of hidden-unit activations from networks trained with neural data $r=0.1$ for the first 10 epochs of training. C) Average variance per unit in the V1-like layer of trained CORNet-Z networks with different weightings, $r$, of neural representations in the cost function. Inset zooms in on the first data points for $r=0, 0.01, 0.1$. It is important to note that the exact outcome of the t-SNE embedding is sensitive to differences in settings.} 
\end{figure}
\subsection{The Details of the ``Teacher" Representation Matter}
Training neural networks to categorize objects, while mimicking the image representations seen in monkey V1, leads to improved object recognition performance (Fig. 2). Is that result specific to the image representations in the monkey brain, or would having \emph{any} arbitrary added RSM constraint in the cost function yield similar results? To answer this question, we performed the same neural network experiment described above (Fig. 2), but using randomly generated matrices in place of the monkey V1 representational similarity matrices. For these experiments, we used the optimal cost function weighting ratio, $r$, found from our experiments with V1 data: $r=0.1$. We repeated these experiments with $r=0.01$ and $r=1.0$ and found similar results (Fig. \ref{required_suppl}). 

We created randomly-generated RSMs in several different ways (see Methods), used them in place of the monkey V1 RSM teacher signal in the training procedure, and compared the trained networks' object categorization performance. Networks trained with the randomly generated RSMs (either with or without matching the mean of the monkey data) under-performed ones trained with real V1 data for the teacher RSM.  Networks trained with shuffled V1 data achieved similar test accuracy to those trained with real V1 data (Fig. 4).  Still, by a small margin, the real V1 data still appears to form the best teacher representation. The teacher RSM may be instructing the DCNN about the distribution of activations (see Sec. 3.2), making the statistics of the data important while not requiring the exact V1 data. These results demonstrate that regularization effects alone -- forcing the network to match an arbitrary RSM -- results in worse performance than when the ``teacher" RSM has statistical properties similar to those seen in V1.  We did not do a full search of drawing from Gaussian distributions with varying mean and standard deviation. Future work could systematically determine at which point the Gaussian-drawn RSM values are no longer close enough to the neural data to be useful. 
\begin{figure}[t]
  \centering
  \includegraphics[trim={0 8cm 0 0}, clip, scale=.48]{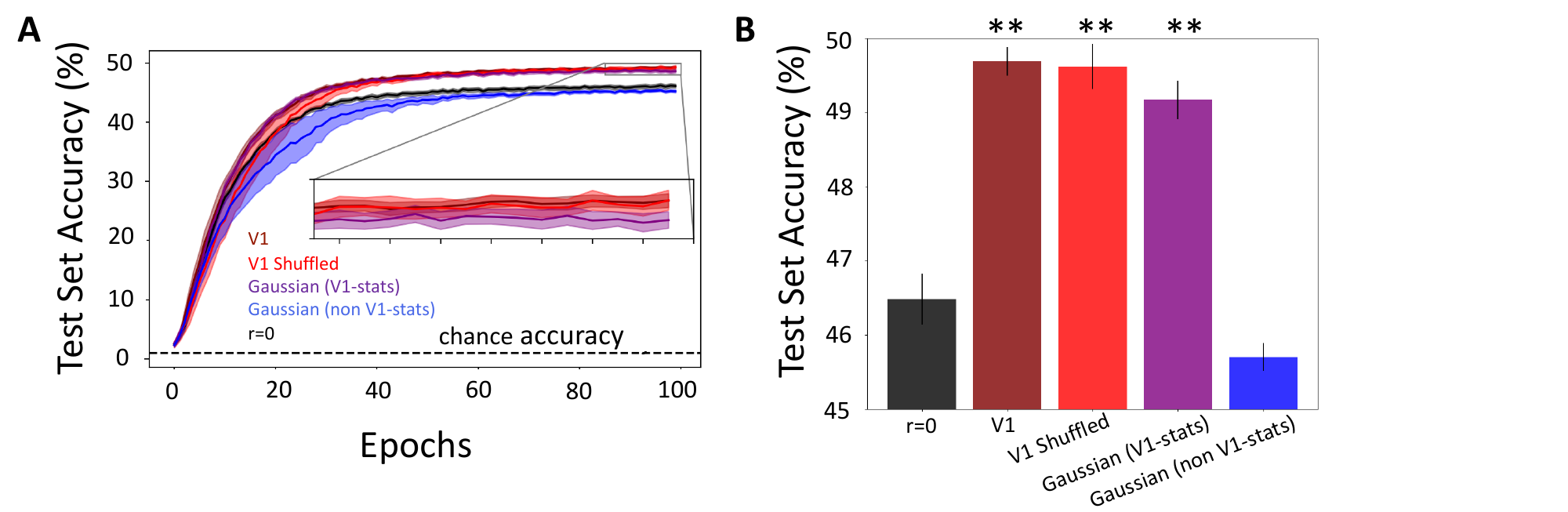}
  \caption{Accuracy in categorizing previously un-seen CIFAR100 images for networks trained with different teacher RSMs: the real monkey V1 RSM (dark red); V1 shuffled RSM (light red); RSM from random Gaussian vectors drawn with the same mean and standard deviation as the V1 data (Gaussian V1-stats in purple); and RSM from random Gaussian vectors drawn with different mean than the neural data (Gaussian non V1-stats in blue). These were all trained with a weighting of $r=0.1$ applied to the representational similarity in the loss function. For comparison, the baseline network (trained with no representational similarity cost) is shown in black. A) Testing accuracy over epochs of training. Shaded areas on plot are +/- SEM over 10 different random initializations of each model. B) Final test accuracy (from A) for RSMs computed from each type of data. Lines on bars are +/- SEM over 10 different random initializations of each model. Double asterisks (**) indicate significantly higher results from no neural data, $r=0$, at p<.001 on a one-tailed t-test. Single asterisks (*) indicate significantly higher results from no neural data, $r=0$, at p<.05 on a one-tailed t-test. }
\end{figure}
\subsection{The Layer of Representation Teaching Matters}
Above, we found that the details of the RSM used as a teacher for training the DCNN matter. This led us to wonder whether it matters \emph{where} in the network that representational similarity cost is applied. To test this, we repeated the experiments from Fig. 2 (training CORNet-Z architecture DCNNs with different weightings for the mismatch between network RSM and monkey V1 RSM), but computed the representation similarity cost for layers other than the V1-identified CORNet-Z layer. The resultant networks all had lower training and testing accuracy than did networks trained with no neural data (Fig. \ref{area_matters}). These results, and those in the preceding section, demonstrate that the performance benefits of the monkey V1 representation teacher require that the monkey V1, or monkey V1-like, representation similarity cost be assigned to the appropriate (early) layer in the DCNN.
\subsection{Networks trained to Mimic Monkey V1 Image Representations Make  More Within Superclass Errors}
We demonstrated that teaching neural networks to respond to images in a more brain-like manner boosts accuracy in categorizing held-out testing data (Fig. 2). All networks, regardless of regularization, still make frequent errors. However, some errors are worse than others. For example, confusing a mouse for a hamster is less severe than confusing a mouse for a skyscraper. This intuition led us to quantify the quality of the errors made by each of our trained networks. For this purpose, we exploited the fact that the 100 classes of labels in the CIFAR100 dataset are grouped into 20 superclasses. One example is ``small mammals", which encompasses mouse, squirrel, rabbit and shrew. 
\begin{figure}[ht!]
  \centering
  \includegraphics[trim={0 4.4cm 0 0}, clip, scale=.45]{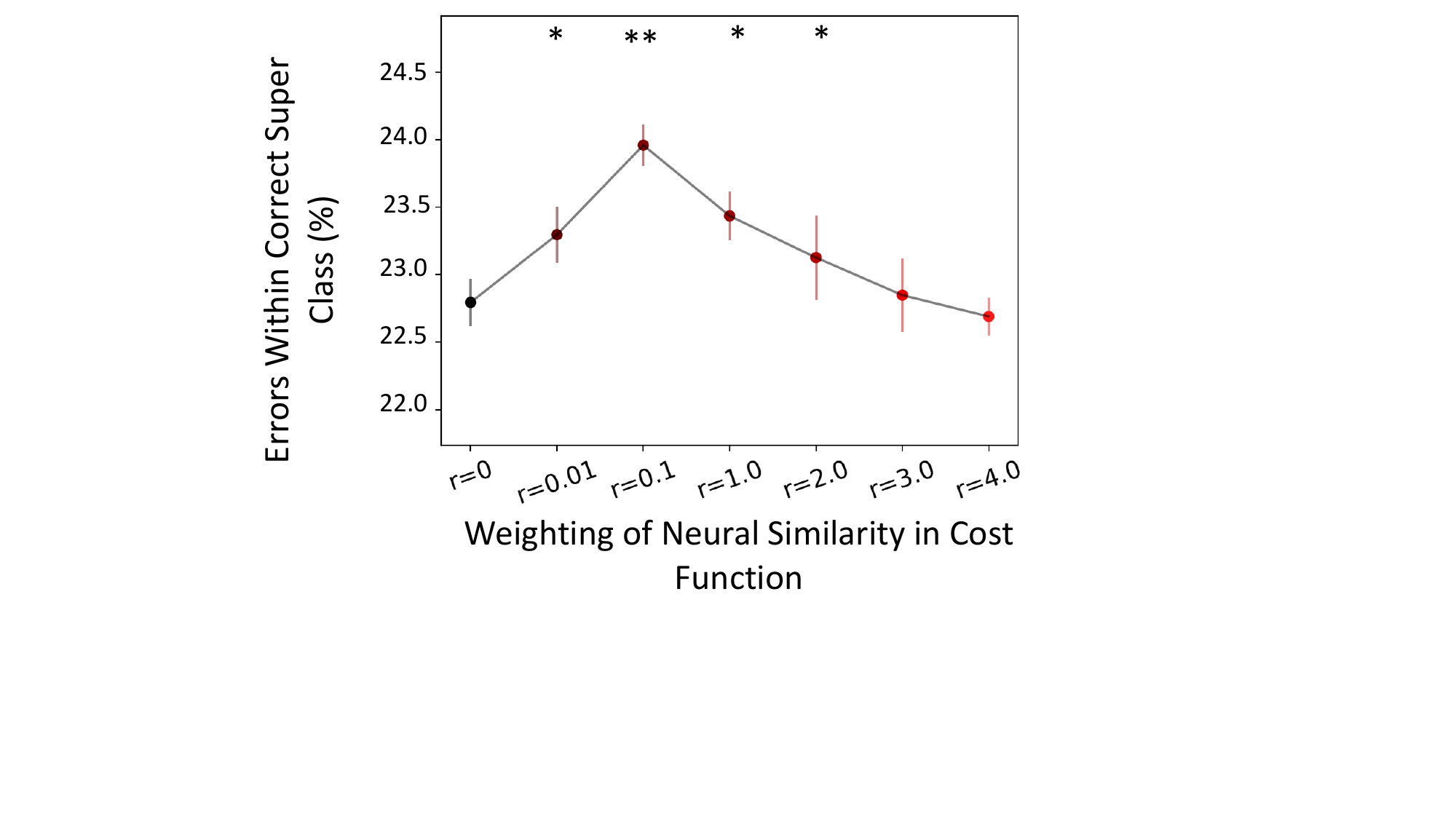}
  \caption{ Percentage of errors within the correct superclass for networks trained with the monkey V1 RSM as a teacher, and various weightings of neural similarity in the cost function (Eq. 1). Error bars are +/- SEM over 10 different random initializations of each model. Double asterisks (**) indicate significantly higher results from no neural data, $r=0$, at p<.001 on a one-tailed t-test. Single asterisks (*) indicate significantly higher results from no neural data, $r=0$, at p<.05 on a one-tailed t-test. }
\end{figure}
We thus asked, for each trained network, what fraction of their categorization errors were within the correct superclass (e.g., confusing a hamster for a mouse) vs in the wrong superclass (e.g., confusing a mouse for a skyscraper). We performed this test on the network trained with the monkey V1 data as a teacher, with the weighting ratio that yielded the best categorization performance ($r=0.1$). For this network, errors were within the correct superclass 23.9\% of the time (Fig. 5, Table 1). For comparison, for the network trained with no neural data, errors were within the correct super class 22.8\% of the time (Fig. 5, Table 1).

Networks trained to categorize images while using monkey V1 as a teacher make fewer categorization errors (Fig. 2). When they do make errors, those errors are more often within the correct superclass and thus likely to be less severe (Fig. 5, Table 1). We find a similar pattern in networks trained with other weighting ratios $r$ (Fig. 5). Moreover, results with randomly generated RSMs mirror their impact on categorization: training with a teacher RSM results both in improved accuracy and an increase in the fraction of errors that are within the correct superclass (Table 1). 

\begin{center}
\captionof{table}{Percentage of errors that are within the correct superclass in the CIFAR100 dataset, for networks trained using different data for the representation similarity regularizer.}
	\begin{tabular}{ |p{4cm}|p{4cm}|p{4cm}|   }
 	\hline
	Weighting of Neural Similarity in Cost Function & Regularization Data & Errors Within Correct Super Class(\%) +/- SEM \\
	 \hline
	 r=0       & None                  			        & 22.8\% +/- 0.002 \\
 	r=0.1    & V1                        			   & 23.9\% +/- 0.002  \\
 	r=0.1    & V1 Shuffled           			   & 23.9\% +/- 0.001 \\
	 r=0.1    & Gaussian (V1-stats) 		   & 23.7\% +/- 0.001 \\
	 r=0.1    & Gaussian (non V1-stats)   & 22.6\% +/- 0.002 \\
 	\hline
	\end{tabular}
\end{center}

\subsection{Networks trained to Mimic Monkey V1 Image Representations are More Robust to Label Corruption}

Given that networks trained using neural data made fewer errors, and the errors they did make were more often within-superclass errors, we hypothesized that the networks would also generalize more robustly in the face of noisy labels. To test that hypothesis, we performed experiments in which some image labels in the training data-set were incorrect. Data-sets will often contain mislabeled images, and ideally computer vision networks would not be heavily affected by these. To generate a corrupted training set, we shuffled labels of 10\% to 50\% of the training data-set, keeping the distribution of classes equal. We only corrupted the training data, and left testing data intact. We then trained CORNet-Z networks with $r=0.1$, and networks without neural data, $r=0$. By the end of the 100 training epochs, networks trained with neural data achieved better testing accuracy than did those without neural data (Fig. 6A). We also found that networks trained with random RSMs drawn from distributions matching neural statistics achieved better testing accuracy than did those trained without neural data (Fig. 6C). The generalization error (training loss - testing loss) was also lower for networks trained with neural data (Fig. 6B). Networks trained with neural data are more robust to mislabeled images than those trained without neural data. 
\begin{figure}[t]
  \centering
  \includegraphics[trim={0 8.2cm 0 0}, clip, scale=.43]{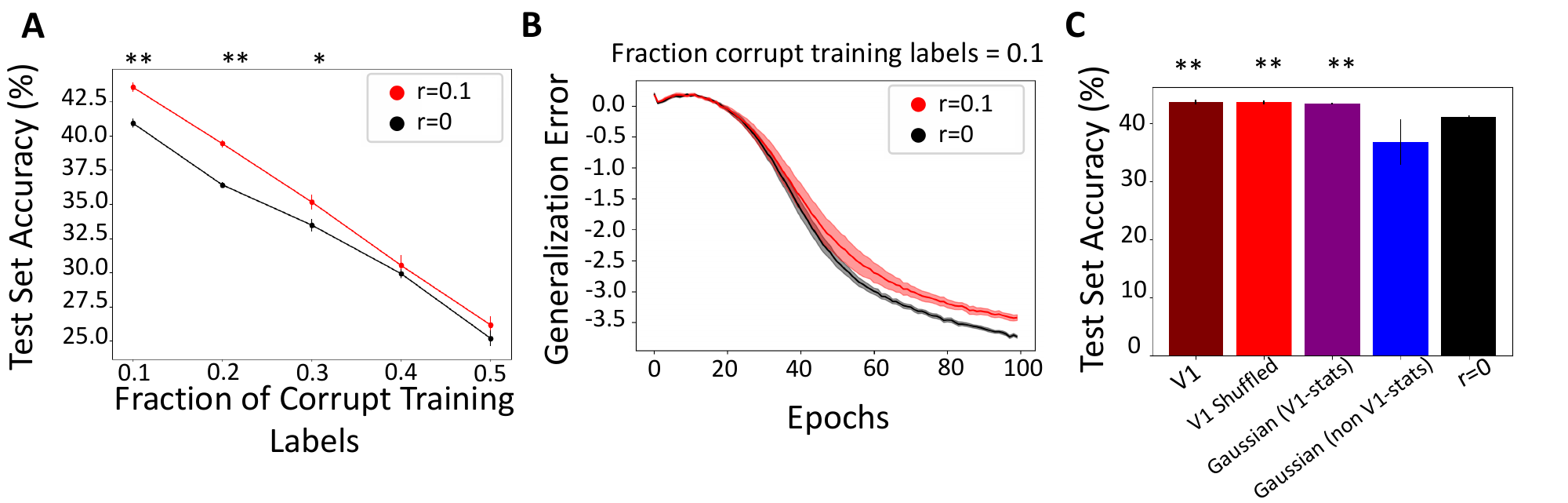}
  \caption{Training CORNet-Z networks with corrupted labels. A) Test set accuracy on networks trained with (red) and without (black) neural data, with different fractions of corrupted training labels. Labels for the test set were not corrupted.  B) Generalization error (training loss - testing loss) on networks trained with (red) and without (black) neural data, and 0.1 fraction corrupt training labels. C) Test set accuracy for networks trained with 0.1 fraction corrupt training labels. The networks were trained with the real monkey V1 RSM (dark red); V1 shuffled RSM (light red); RSM from random Gaussian vectors drawn with the same mean and standard deviation as the V1 data (Gaussian V1-stats in purple); RSM from random Gaussian vectors drawn with different mean than the neural data (Gaussian non V1-stats in blue); and no neural data (black).  Error bars in A, C and shaded areas in B are +/- SEM over 10 different random initializations of each model. For A, C: Double asterisks (**) indicate significantly different results from no neural data, $r=0$, at p < 0.001 on a one-tailed t-test. Single asterisks (*) indicate significantly higher results from networks trained with no neural data, $r=0$, at p < 0.05 on a one-tailed t-test. }
\end{figure}
\subsection{Similar Results are Obtained with the Larger VGG-16 Network} \label{VGG}
All of the above experiments were run with the relatively small CORNet-Z network. This led us to wonder whether monkey V1 data could serve as a similarly effective teacher for deeper neural networks that achieve higher baseline performance. To answer this question, we repeated our above experiments with the VGG-16 architecture \cite{VGG16} instead of CORNet-Z. We applied the monkey V1 representational similarity cost at the third convolutional layer of the VGG-16 (see Methods). We found that, with the VGG-16 architecture, the monkey V1 teacher signal improves categorization performance, increases robustness to corrupted labels in the training set, and reduces generalization error, similar to what was observed with the smaller CORNet-Z architecture (Fig. 7). 
\begin{figure}[t]
  \centering
  \includegraphics[trim={0 0cm 0 0}, clip, scale=.47]{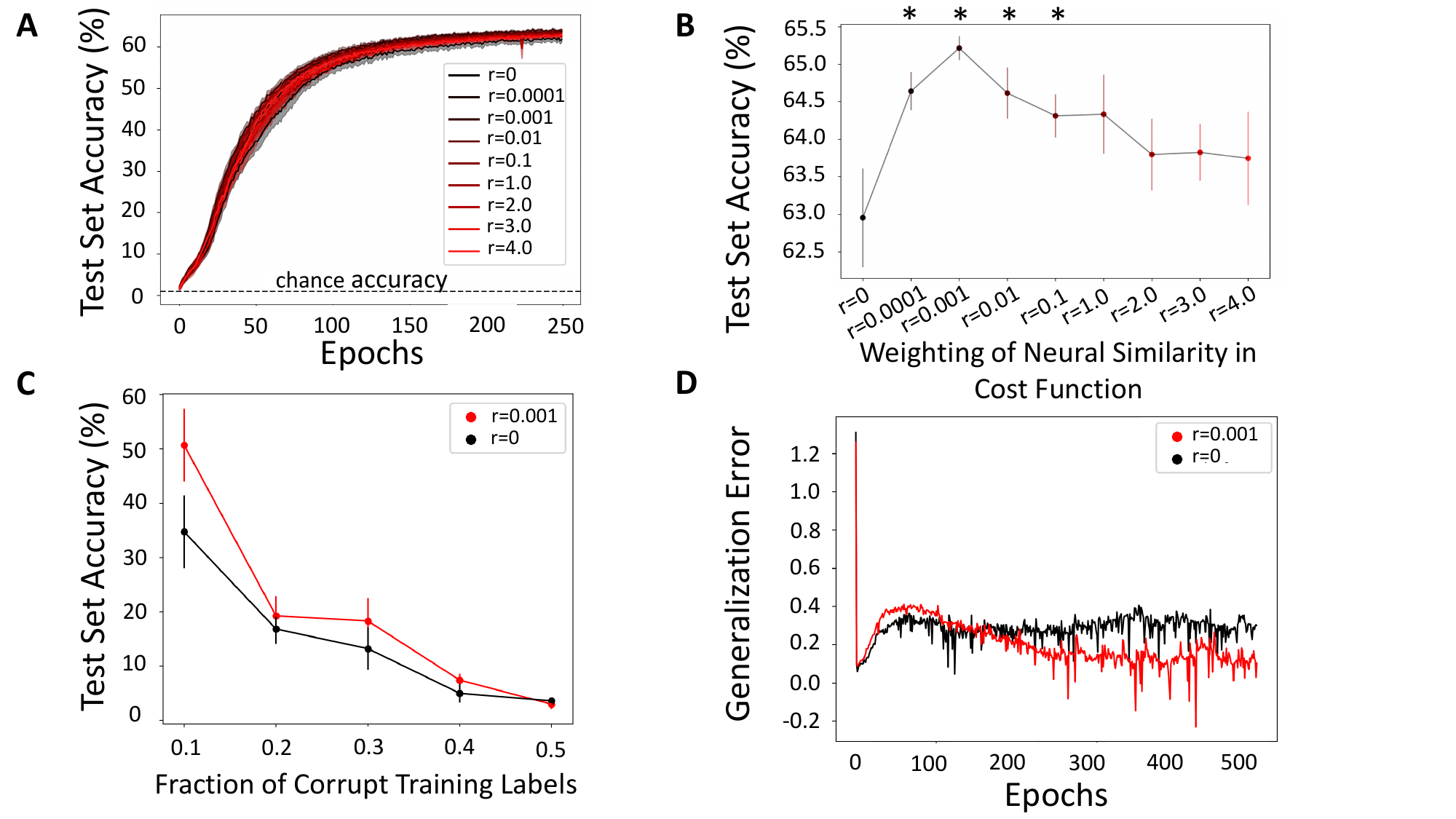}
  \caption{Accuracy in categorizing previously-unseen CIFAR100 images for the VGG-16 architecture trained on the
composite task using monkey V1 representational similarity. A) Testing accuracy for each epoch of training for several values of $r$. Chance accuracy is indicated by the dashed black line. B) Test accuracy plotted (same as in A) as a function of the emphasis on neural representation similarity (controlled by $r$). Single asterisks (*) indicate significantly higher results from no neural data, $r=0$, at p < 0.05 on a one-tailed t-test. C) Test set accuracy on networks trained with (red) and without (black) neural data, with different fractions of corrupted labels in the training set. Labels for the test set were not corrupted.  Shaded areas and error bars on the plots are +/- SEM over 10 different random initializations of each model. D) Generalization error (training loss - testing loss) on networks trained with (red) and without (black) neural data and 0.1 fraction corrupted training labels.  }
\end{figure}
\section{Discussion}
Training the early layers of convolutional neural networks to mimic the image representations from monkey V1 improves those networks' ability to categorize previously-unseen images. Moreover, networks trained using monkey V1 as a representation ``teacher" made errors that were more often within the correct superclass than that did networks without the ``teacher'' signal. While the performance gains were modest, they were remarkably robust: we observed similar performance gains on large and small architectures (CORNet-Z and VGG-16), using different optimizers (Adam and gradient descent), using different activation functions (ELU and ReLu). This robustness suggests that our method could generalize well to other networks.

We focused on feedforward convolutional neural networks, however, the mammalian visual system (MVS) has substantial recurrent connectivity  \cite{VictorAFLammeHansSuper1998}. Interesting future work would be to implement our same approach in recurrent convolutional neural networks which may demonstrate more performance improvement from training with neural data \cite{Spoerer2017}. We also focused on one area of adapting the loss function to incorporate both object recognition and matching neural data. Interesting related work to adapting the loss function introduced using robustness as a continuous parameter to the loss function \cite{Barron2019}. Important future work could consider alternative approaches to adapting the loss function to the specific task, rather than using a more generic one like cross entropy. 

Our experiments indicate that the details of the ``teacher" representation are important: making the earlier layers of the neural network have mimic monkey V1 activation patterns is helpful for categorization performance. Interestingly, we found that  similar results were obtained with randomized data as were obtained using the real monkey V1 data, so long as those randomized data matched the statistics of monkey V1 neural responses. Thus, the ?teacher? RSM is likely teaching the network about the response distributions that are most useful for the categorization task, pointing to a potential role for RSM regularization to succeed even in cases where no monkey brain data is explicitly used. (E.g., the network could be trained to mimic the known statistical properties of neural responses, even if no brain data is explicitly used in the training process).

It is important to note that the data we used here were from anesthetized monkeys, who were shown the images for very brief periods ($100$ ms). Moreover, with the recording method used (Utah arrays implanted in V1), only O(10) neurons could be simultaneously observed. Despite the caveats of our dataset, we found that the statistically closer the teacher representation was to the real neural data, the better the trained neural network did at object recognition. A key future direction -- that we are actively pursuing -- is to repeat the experiments here, but with data from awake animals, doing vision-based tasks, in which a larger fraction of V1 is observed. We anticipate that larger performance gains could be achieved in those experiments.

We emphasize that our goal here was not to achieve state-of-the-art classification performance, but rather to determine whether and how we could use the brain as a teacher for training artificial neural networks to perform object recognition tasks. We tested this with over a wide range of network conditions, and demonstrated that using the brain as a teacher signal results in robust performance gains. Our expectation is that future work -- by ourselves and others --  could apply this technique to develop better object recognition systems.

In addition to the V1 data, which were the main focus of this paper, we also tried training our networks with V4 and IT recorded data from macaque monkeys \cite{Majaj2015}. Using the CORNet-Z architecture, we experimented with forcing the deeper layers of the network to match representations from those recordings. This resulted in lower testing accuracy than the baseline CORNet-Z model. However, we cannot conclusively state that V4 or IT data would not be helpful in training object recognition networks. The decrease in accuracy could be due to not finding the optimal combination of how many epochs for which to train the network on the joint loss function, and how much to emphasize matching the neural representations. It could also be because the data normalization done within the lab that collected the V4 and IT \ data \cite{Majaj2015} was different from the lab from which our V1 data came \cite{Coen-cagli2015}. However, it remains possible that V1 data may be actually more useful for training DCNNs because the specificity of neural responses increases in the deeper layers of the mammalian visual system: the V4 and IT data may not include recordings from neurons that would be useful in our particular object recognition task, whereas V1 neurons carry more generalist representations. Finally, features in DCNNs are most transferable between tasks in the early layers, and become much less so in deeper layers meant to represent V4 and IT \cite{Yosinski2014}. Because the images shown to the monkeys in our experiments differed from those in the CIFAR100 categorization task, this further suggests that data from lower visual areas, like V1, could be the most useful brain data for training object recognition networks.

Our work provides solid evidence that neural recordings can be used to ``teach'' machine learned models to better categorize the objects in images.  This same concept could be applied in many application areas where neural networks are trained to perform something humans or animals can do easily (e.g. language understanding, game playing).  Furthermore, we have provided evidence that we do not necessarily need to fully understand how the brain works in order to impart that operational skill onto an artificial neural network.

\section*{Appendix}
\setcounter{figure}{0} \renewcommand{\thefigure}{A.\arabic{figure}}

\subsection*{A.1 Training With Static Cost Weightings} \label{staticSec}
\begin{figure}[H]
 \centering
 \includegraphics[trim={0 8.1cm 0 0}, clip, scale=.45]{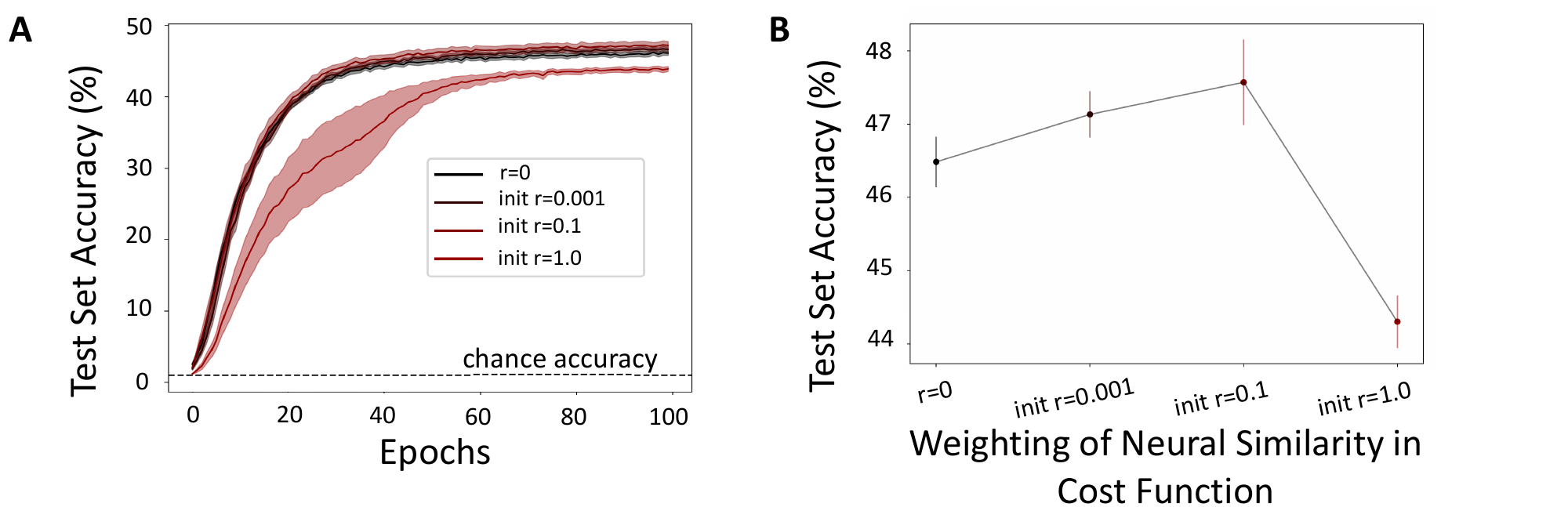}
\caption{Accuracy in categorizing previously-unseen CIFAR100 images for the CORNet-Z architecture trained with static $\lambda$ values applied to monkey V1 representation similarity for all 100 epochs of training. The test-set accuracy is plotted at each epoch during training. Chance accuracy is indicated by the dashed black line. Shaded areas are +/- SEM over 10 different random initializations of each model.  B) Test-set accuracy for previously-unseen CIFAR100 images at the end of training, as a function of the initial static weight, $init$ $r$, given to neural representational similarity in the cost function. }
\label{static_ratios}
\end{figure}

In the experiments in the main paper, we adjusted the weighting of the representation similarity cost in the loss function ($\lambda$) during training, so as to keep a constant ratio ($r$) between the cross-entropy cost and the representation similarity cost. We also applied the representation similarity cost only for the first 10 epochs of training. A natural question is whether similar results would be achieved with a static $\lambda$ value, or with the representation similarity cost applied for longer duration during training. To answer these questions, we repeated our experiments from Fig. 2, with static $\lambda$ values, and with $\lambda \ne 0$ for the first 100 epochs of training (instead of 10). We found that this method did not work as well as the one described in Fig. 2: to see this, compare the accuracy values in Fig. 2 to those in Fig. \ref{static_ratios}. 

\subsection*{A.2 Training Over Longer Timescales} \label{timescalesSec}
\begin{figure}[H]
 \centering
 \includegraphics[trim={0 7cm 0 0}, clip, scale=.45]{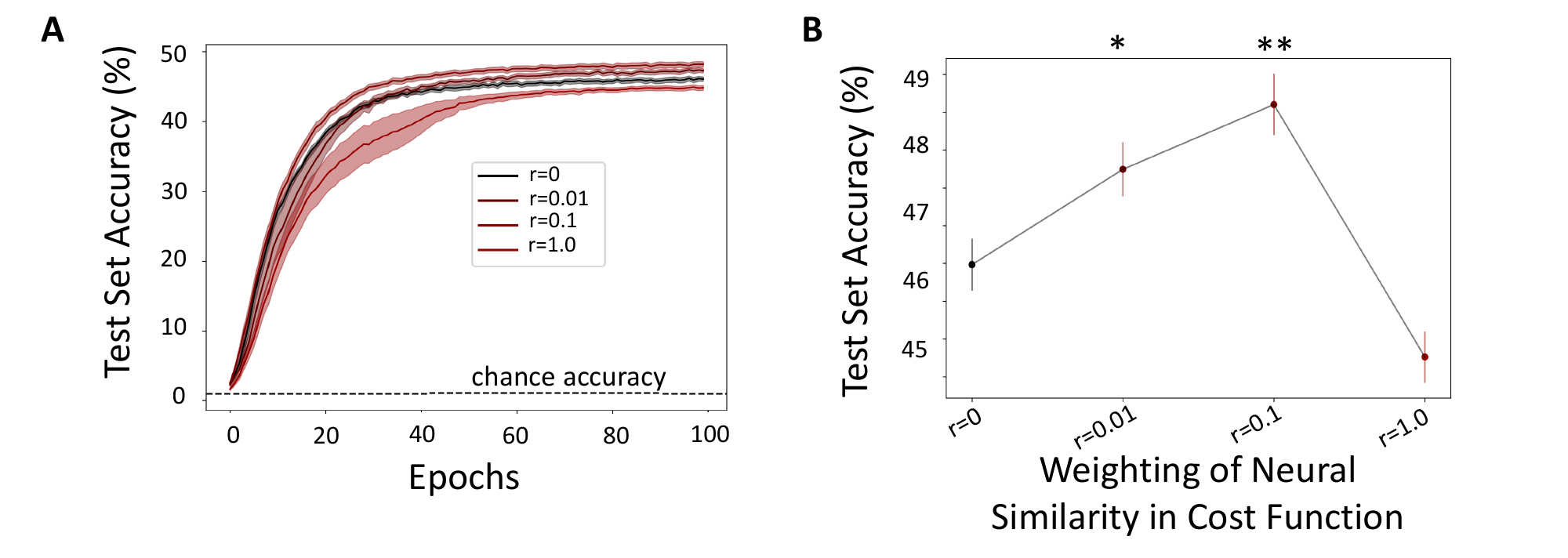}
\caption{Training with representation similarity cost at all epochs. (Similar to Fig. 2 but with representation similarity cost applied for all 100 epochs). A) Testing accuracy for each epoch of training with static weighting ratio, $r$. B) Test accuracy (same as in A) vs the weight of the emphasis on neural representation similarity.  Double asterisks (**) indicate significantly higher results from no neural data, $r=0$, at p<.001 on a one-tailed t-test. Single asterisks (*) indicate significantly higher results from no neural data, $r=0$, at p<.05 on a one-tailed t-test. Chance accuracy is indicated by dashed black line in A. Shaded areas and vertical bars on plot are +/- SEM over 10 different random initializations of each model.}
\label{allepochs}
\end{figure}

In the main paper, we used neural data as a teacher signal for only during the first 10 epochs of the training procedure: after the 10th epoch, the weighting parameter, $r$, was set to zero. This led us to wonder whether larger performance benefits might be obtained by extending the time window over which the neural data contributed to the training. To answer this question, we trained networks with neural data for all 100 epochs, instead of the first 10, and found a similar boost in testing accuracy for networks trained with neural data (Figs. \ref{allepochs} A, B). This demonstrates that we do not need to run the regularization step for longer than the first 10 epochs. 

\subsection*{A.3 The Details of the Teacher Representation Matter}
In section 3.3, we compared neural networks trained with different teacher RSMs. For those experiments, we used the optimal cost function weighting ratio, $r$, found from our experiments with the real V1 RSM: $r=0.1$. Those experiments showed that, as the teacher RSM better approximated the monkey V1 RSM, the trained neural network achieved better performance. This left open the question of whether training the networks with our randomized ``control" RSMs, with different weighting ratios (i.e., not $r=0.1$) would lead to better performance.

\begin{figure}[H]
\centering
 \includegraphics[trim={0 0cm 0 0}, clip, scale=.46]{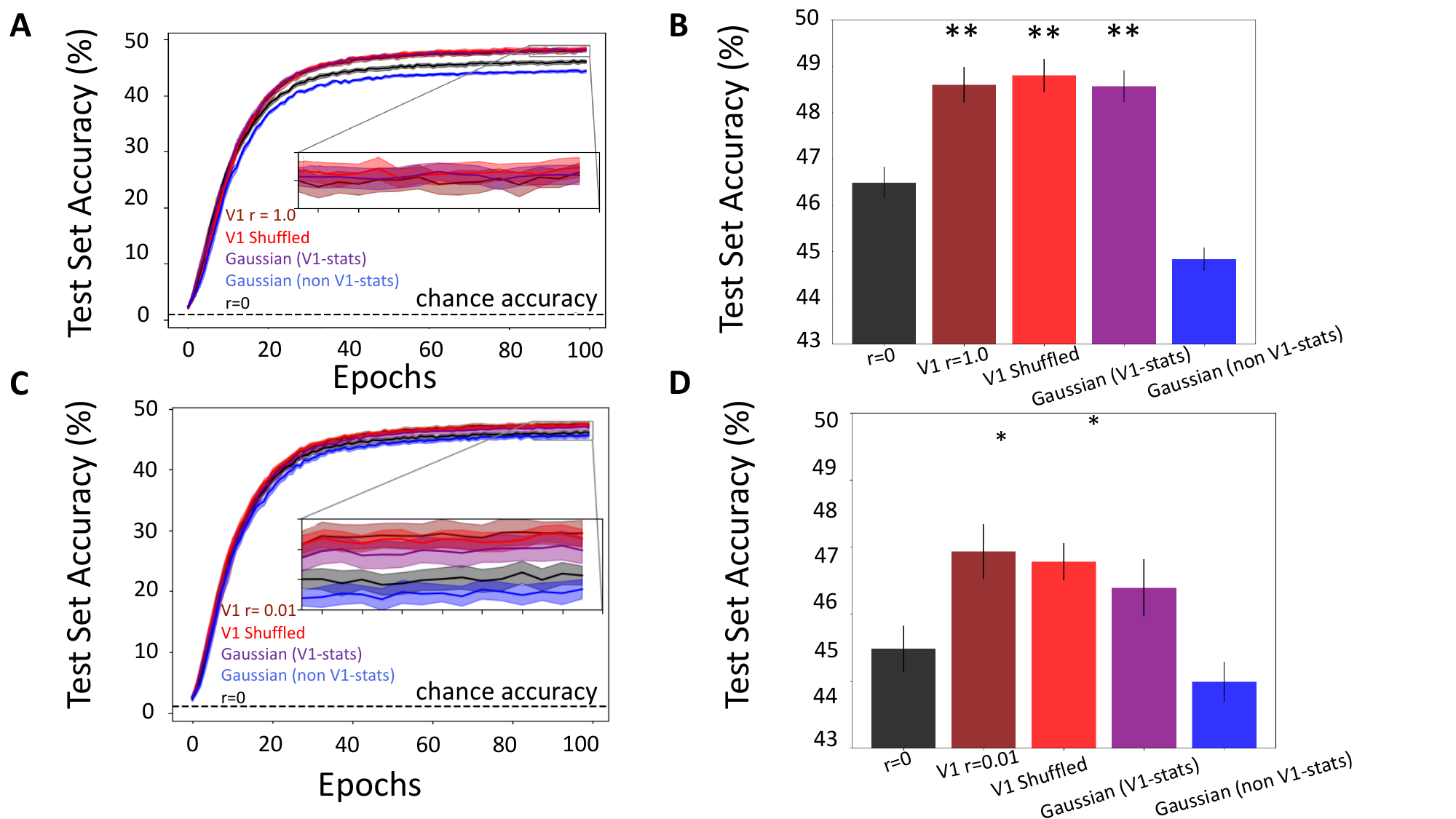}
 \caption{Accuracy in categorizing previously un-seen CIFAR100 images for networks trained with different teacher RSMs: the real monkey V1 RSM (dark red); V1 shuffled RSM (light red); RSM from random Gaussian vectors drawn with the same mean and standard deviation as the V1 data (Gaussian V1-stats in purple); and RSM from random Gaussian vectors drawn with different mean than the neural data (Gaussian non V1-stats in blue). These were all trained with a weighting of  $r=1.0$ (top panel) and $r=0.01$ (bottom panel) applied to the representational similarity in the loss function. For comparison, the baseline network (trained with no representational similarity cost) is shown in black. A) and C) Testing accuracy over epochs of training. Shaded areas on plot are +/- SEM over 10 different random initializations of each model. B) and D) Test accuracy plotted (same as in A) by type of data used in forming the teacher RSM. Lines on bars are +/- SEM over 10 different random initializations of each model. Double asterisks (**) indicate significantly higher results from no neural data, $r=0$, at p<.001 on a one-tailed t-test. Single asterisks (*) indicate significantly higher results from no neural data, $r=0$, at p<.05 on a one-tailed t-test. }
\label{required_suppl}
\end{figure}

To address this question, we repeated our experiments with other weighting ratios: $r=0.01$ and $r=1.0$. We found that, for all RSMs, the best performance was found with $r=0.01$ (the one used in Sec. 3.3): compare values in Fig.  4 and Fig. \ref{required_suppl}.

\subsection*{A.4 The Layer of Representation Teaching Matters} \label{timescalesSec}
\begin{figure}[H]
 \centering
 \includegraphics[trim={0 8cm 0 0}, clip, scale=.45]{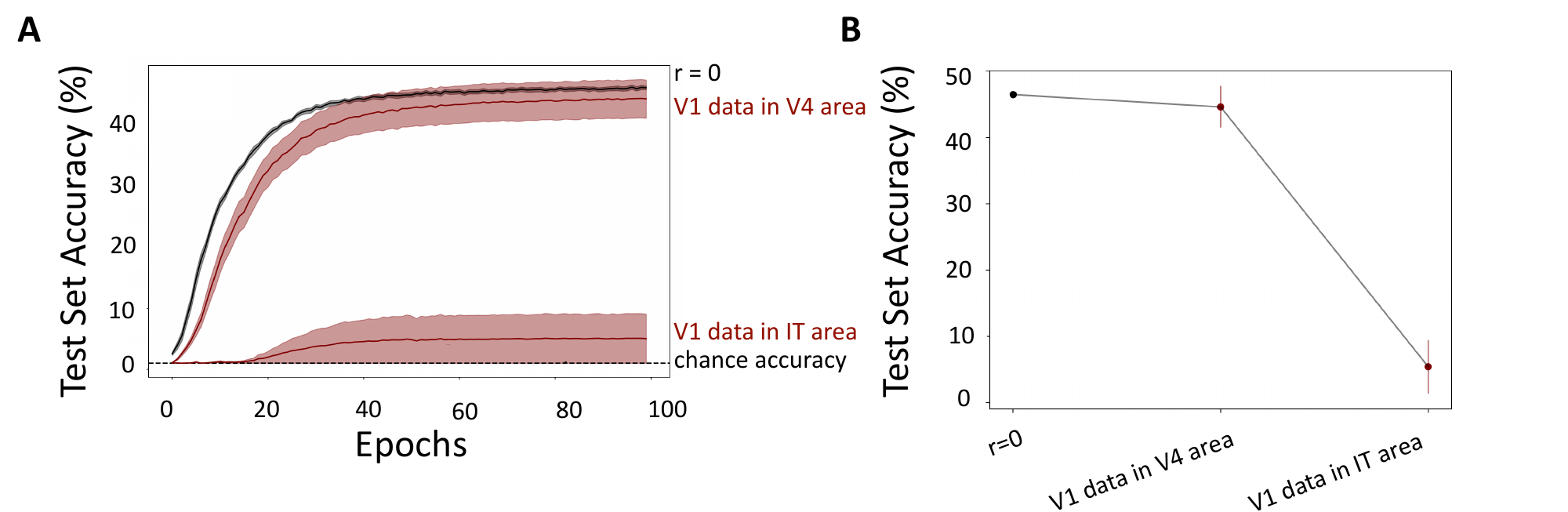}
\caption{Accuracy in categorizing previously-unseen CIFAR100 images for the CORNet-Z architecture trained with V1 data in the V4-like area of the CORNet-Z model (top red line) and V1 data in IT-like area of the CORNet-Z model (bottom red line) with $r=0.1$. A) Test-set accuracy at each epoch during training. Chance accuracy is indicated by dashed black line. Shaded areas are +/- SEM over 10 different random initializations of each model.  B) Test-set accuracy for previously-unseen CIFAR100 images at the end of training, as a function of neural representation area and weighting.}
\label{area_matters}
\end{figure}

We trained CORNet-Z architecture DCNNs on our composite task, and used the monkey V1 RSM as a teacher for either the V4, or the IT-like areas of CORNet-Z (see Fig. 1 for correspondence between layers). Training the V4-like CORNet-Z layer to mimic monkey V1 led to slightly worse results than training with no teacher RSM ($r=0$ in black) (Fig. \ref{area_matters}). Training the IT-like layer of CORNet-Z to mimic monkey V1 led to much lower performance (Fig. \ref{area_matters}). This demonstrates that the use of V1 representations as a teacher for the V1-like layer of the neural network is important for performance improvements. 

\subsection*{A.5 Natural Images}
\begin{figure}[H]
\centering
 \includegraphics[trim={0 10cm 0 0}, clip, scale=.48]{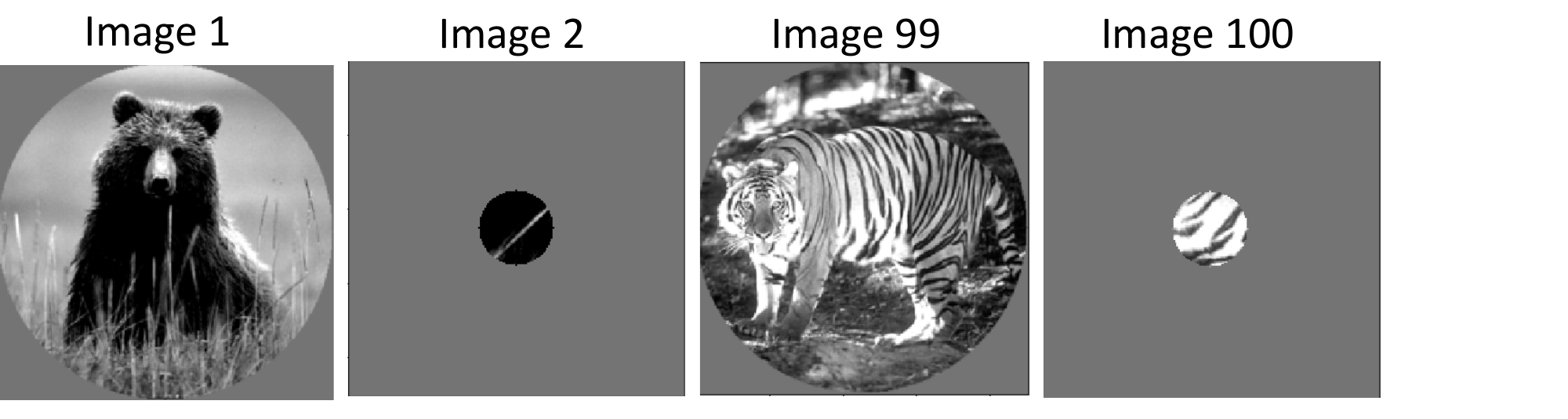}
\caption{A sample of 4 out of the 270 natural images shown to the anesthetized monkeys for our V1 teacher signal \cite{Coen-cagli2015}. The images are of natural scenes (Images 1 and 99), or small sections cropped out of natural scenes (Images 2 and 100)}. 
\label{sample_figs}
\end{figure}

Our monkey V1 teacher signal is from publicly-available multielectrode recordings from anesthetized monkeys presented with a series of  natural images while experimenters recorded the spiking activity of neurons in primary visual cortex (V1) with a multielectrode array \cite{Coen-cagli2015}. The monkeys were shown 270 static natural images as well as various static grating images for 100 ms presentations. Example images are in Fig \ref{sample_figs}).

\subsection*{A.6 Distribution of Test Accuracy}
\begin{figure}[H]
\centering
 \includegraphics[trim={0 6.5cm 0 0}, clip, scale=.5]{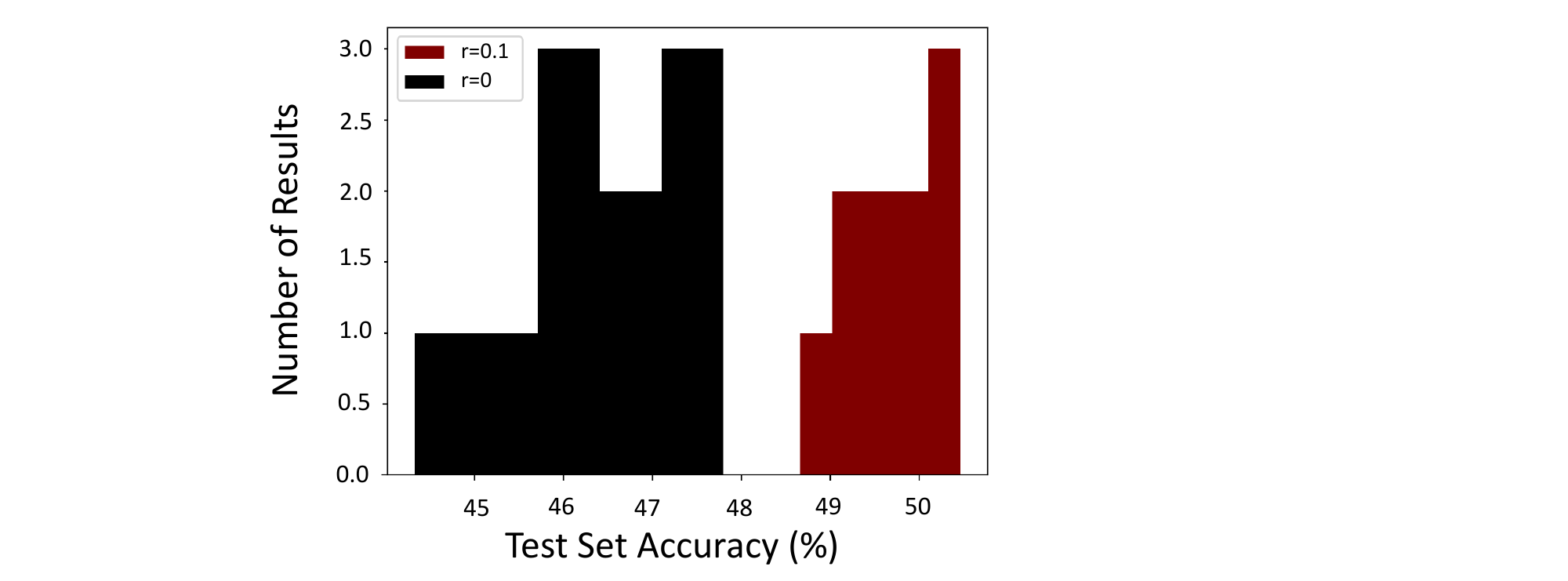}
\caption{A histogram of the accuracy in categorizing previously-unseen CIFAR100 images for the CORNet-Z architecture of networks trained with no neural data, $r=0$ (black), and networks trained with neural data, $r=0.1$ (red) for the first 10 epochs of training. These results are the same as those found in Fig. 2. }
\label{hist}
\end{figure}

To better visualize the range of test set accuracy achieved on networks trained with and without neural data, we plotted the histogram of maximum accuracy found across the 10 random initializations of networks trained with the optimal ratio of neural data, $r=0.1$, and without neural data, $r=0$. (Fig. \ref{hist}). 

\subsection*{A.7 ELU Activation Functions} 
To demonstrate that our improved performance with neural data is not specific to the activation function, we also experimented with using Exponential Linear Units (ELUs) \cite{Clevert2015}. We trained 10 randomly initialized networks with the optimal ratio of neural data found with the ReLU activation function, $r=0.1$, and without neural data, $r=0$, with the same architecture setup as described in Fig. 1, except we used the ELU activation function instead of the ReLU activation function.  We found that networks trained with neural data outperform those trained without neural data, and that performance with ReLU activations outperform networks trained with ELU activations for networks with and without neural data.  Networks with ELU activation and neural data ratio $r=0.1$ (red) had an average max accuracy of $47\% +/- 0.002$. Networks with ELU activations and no neural data $r=0$ (black) had an average max accuracy of $41.8 +/- 0.45$. The results with and without neural data is not significantly different at p<0.05 on a one-tailed t-test.

\begin{figure}[H]
\centering
 \includegraphics[trim={0 8cm 0 0}, clip, scale=.7]{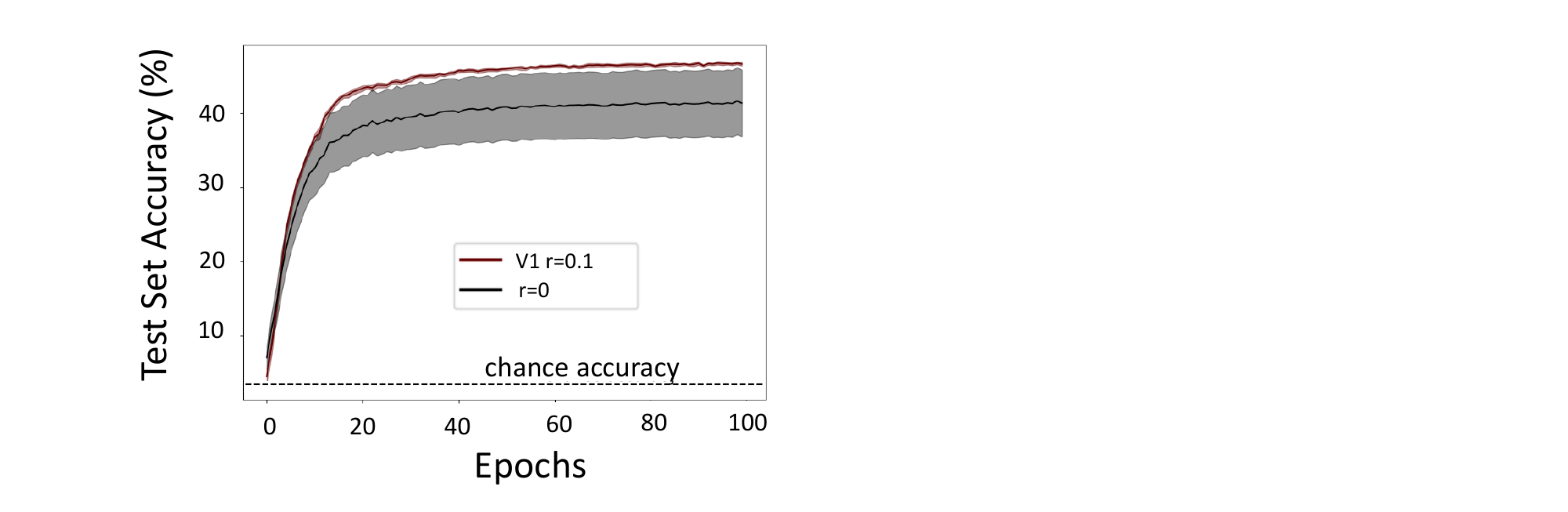}
\caption{Accuracy in categorizing previously-unseen CIFAR100 images for the CORNet-Z architecture of networks trained with no neural data, $r=0$ (black), and networks trained with neural data,  $r=0.1$ (red) for the first 10 epochs of training. Networks were trained with the ELU activation functions. Chance accuracy is indicated by the dashed black line. Shaded areas are +/- SEM over 10 different random initializations of each model. }
\label{elu}
\end{figure}

\subsection*{A.8 Adam Optimizer} 
To demonstrate that our improved performance with neural data is not specific to the optimizer or learning rate schedule, we also experimented with the Adam optimizer \cite{Kingma2015}, which incorporates an automated learning rate scheduler.  We trained 10 randomly initialized networks with the optimal ratio of neural data, $r=0.1$, and without neural data, $r=0$, with the same architecture setup as described in Fig. 1, except that we used the Adam optimizer instead of the gradient descent optimizer. We found that the Adam optimizer improved the baseline performance for CORNet-Z trained without neural data from $46.484\% +/- 0.3434$ to $49.167\% +/- 0.0009$. We found similar results for networks trained with neural data with the Adam optimizer and the gradient descent optimizer. Networks trained with neural data and gradient descent had a max average accuracy of $49.689\% +/- 0.1922$ versus $50.062\% +/- 0.0017$ for networks trained with the Adam optimizer. Networks trained with neural data out-perform networks trained without neural data. Networks trained with neural data are significantly different at $p<0.001$ one a one-tailed t-test.

\begin{figure}[H]
\centering
 \includegraphics[trim={0 8cm 0 0}, clip, scale=.65]{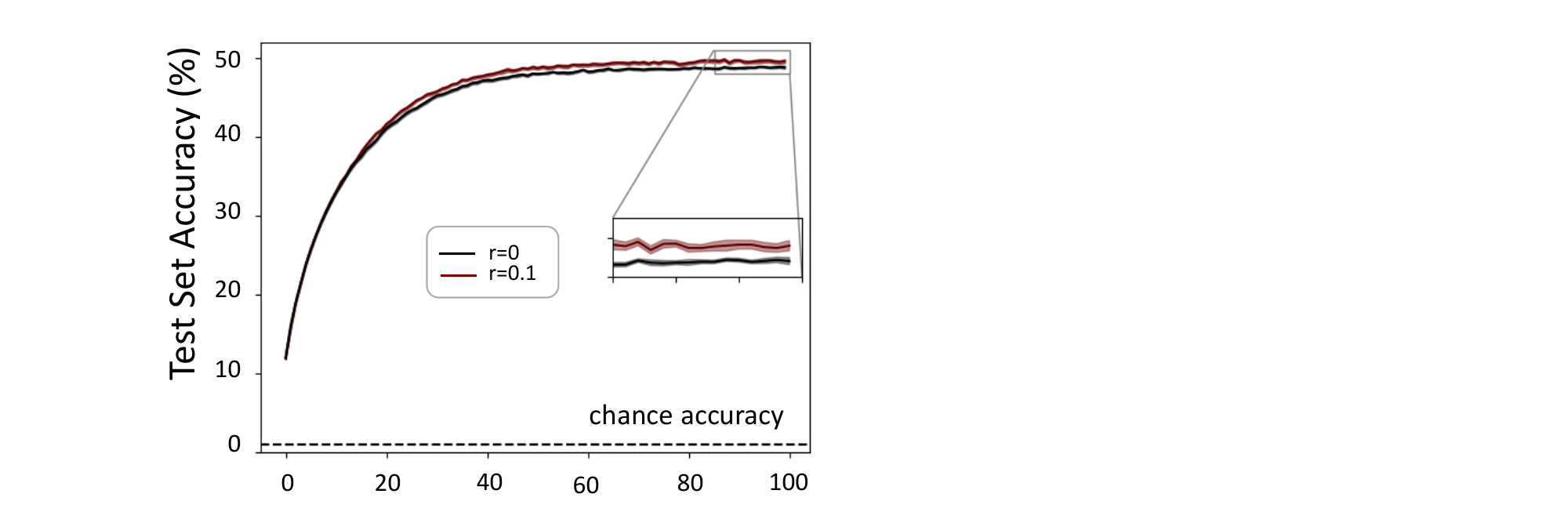}
\caption{Accuracy in categorizing previously-unseen CIFAR100 images for the CORNet-Z architecture of networks trained with no neural data, $r=0$ (black), and networks trained with neural data, $r=0.1$ (red) for the first 10 epochs of training. Networks were trained with the Adam optimizer. Chance accuracy is indicated by the dashed black line.  Shaded areas are +/- SEM over 10 different random initializations of each model.  }
\label{adam}
\end{figure}

\subsubsection*{Acknowledgments}
JZ is an Associate Fellow of CIFAR, in the Learning in Machines and Brains Program. JZ further acknowledges the following funding sources: Sloan Fellowship, Canada Research Chairs Program, and Natural Science and Engineering Research Council of Canada (NSERC). CF was supported by an NSF Graduate Research Fellowship, Award $\# 1553798$.  AF is a Fellow of CIFAR program for Learning in Machines and Brains, and holds a Canada CIFAR AI Chair.  AF and HX are funded through CIFAR and an NSERC Discovery Grant. 

{\small
\bibliography{references}}
\bibliographystyle{plain}

\end{document}